# A Variance Maximization Criterion for Active Learning

**Yazhou Yang**                                                Y.YANG-4@TUDELFT.NL
**Marco Loog**                                                M.LOOG@TUDELFT.NL
Pattern Recognition Laboratory, Delft University of Technology, Van Mourik Broekmanweg 6, 2628 XE Delft, The Netherlands

## Abstract

Active learning aims to train a classifier as fast as possible with as few labels as possible. The core element in virtually any active learning strategy is the criterion that measures the usefulness of the unlabeled data based on which new points to be labeled are picked. We propose a novel approach which we refer to as maximizing variance for active learning or MVAL for short. MVAL measures the value of unlabeled instances by evaluating the rate of change of output variables caused by changes in the next sample to be queried and its potential labelling. In a sense, this criterion measures how unstable the classifier's output is for the unlabeled data points under perturbations of the training data. MVAL maintains, what we refer to as, retraining information matrices to keep track of these output scores and exploits two kinds of variance to measure the informativeness and representativeness, respectively. By fusing these variances, MVAL is able to select the instances which are both informative and representative. We employ our technique both in combination with logistic regression and support vector machines and demonstrate that MVAL achieves state-of-the-art performance in experiments on a large number of standard benchmark datasets.

## 1. Introduction

In many real-world applications of classification problems, we face the problem that obtaining labels is more difficult than collecting input data: we can easily acquire a large amount of such input data, but labeling these instances is quite burdensome, time-consuming, or expensive (Settles, 2010). For a large part, this is because of the heavy involvement of human supervision during the labeling process. For example, a hospital produces large amounts of digital images every day, but when categorizing these med-ical images one often needs to rely on medical doctors with a particular, and therefore expensive, expertise. Hence, it is essential to reduce the need for human annotation, bringing down cost by labeling fewer yet more informative samples. The problem studied in active learning is how to select the most valuable subset and how to measure the value of individual instances or collections of these.

In this work, we focus on, what we refer to as, retraining-based active learning in which one measures the usefulness of particular instances based on all the possible models that are obtained by adding the instances to the labeled dataset and retraining the classifier with the different labels possible (Roy & Mccallum, 2001; Schein & Ungar, 2007; Tong & Koller, 2002). This means that with $n$ unlabeled points and $k$ different classes to choose from, we train $nk$ different classifiers. The key idea behind this is that the value of an unlabeled instance can be estimated by the change it brings to the model when it is queried and used to retrain the model.

Here we propose a new retraining-based active learning method: maximizing variance for active learning (MVAL). Our method selects the instances with maximum retraining variance. This variance stems from the variation presented in the next sample to query and the possible labels those samples can have. The idea is that if the output of an instance changes dramatically, it means that this instance is very susceptible to the variations of input training data. On the other hand, if an instance's output does not vary much, this indicates that the current classifier is very certain about it. A sample with the largest changes in output value is most uncertain and this rate of change can be naturally measured by the variance. Thus, the larger the variance of the output of an unlabeled instance, the higher the uncertainty it has. We propose to keep track of the estimated probability (or decision output) of each unlabeled instance during the retraining procedure. The recorded information is utilized to produce so-called retraining information matrices (RIMs), which are used to calculate the variances for all unlabeled samples. More specifically, two different kinds of variance are computed to measure the informativeness and representativeness. By selecting the instances with maximum variance, MVAL is able to query instances that are both informative and representative. Furthermore, MVAL can be incorporated with both probabilistic and non-probabilistic classifiers, such as logistic regression, Naive Bayes, support vector machines and least squares classifier. In this paper, we construct the experiments of MVAL with logistic regression and support vector machines.

The remainder is organized as follows. Section 2 reviews related work, focussing on retraining-based active learning algorithms. The proposed method is presented in detail in Section 3, followed by an extension of the proposed method to multiclass classification problems in Section 4. Section 5 and Section 6 report the experimental results on binary and multi-class classification problems, respectively. Finally, we conclude this paper in Section 7.

## 2. Related Work

In the past decades, various active learning algorithms, based on many different selection criteria, have been proposed. These approaches rely on different heuristics. We can roughly divide these heuristics into two categories: informativeness and representativeness. Informativeness estimates the ability of an instance in decreasing the uncertainty of a statistical model, while representativeness indicates whether a sample is representative of the underlying distribution (Settles, 2010). For example, query-by-committee (Seung et al., 1992), uncertainty sampling (Lewis & Gale, 1994; Tong & Koller, 2002; Wang et al., 2011), error reduction (Guo & Greiner, 2007; Roy & Mccallum, 2001), model change (Cai et al., 2014; Freytag et al., 2014; Kading et al., 2015; Settles et al., 2008), expected variance reduction (Schein & Ungar, 2007) belong to the informativeness category, but each of them has its own criterion of informativeness. Clustering-based approaches (Nguyen & Smeulders, 2004; Saito et al., 2015; Xu et al., 2003) and variance minimization methods (Ji & Han, 2012; Lu et al., 2011; Ma et al., 2013; Yu et al., 2006) are included in the representativeness group. There are also methods that try to combine the two criteria, such as min-max view active learning (Huang et al., 2014), density or diversity weighted methods (Brinker, 2003; Liu et al., 2008; Settles & Craven, 2008; Yang et al., 2015; Zhu et al., 2010) and multi-criteria fusion (Du et al., 2017; Wang & Kwong, 2014; Wang et al., 2016; Wang & Ye, 2013).

The framework of retraining-based active learning, which our method is also an instantiation of, was first proposed by Roy & Mccallum (2001) to perform so-called expected error reduction (EER for short). Tong & Koller (2002) used a retraining approach in combination with SVMs to find instances that, after labeling, approximately halve the ver-sion space. A series of active learning methods which propose a scheme similar to EER, but with somewhat different motivations, were put forward in (Evans et al., 2015; Freytag et al., 2014; Guo & Greiner, 2007; Schein & Ungar, 2007). All in all, retraining-based active learners can be roughly divided into four categories: error reduction (Guo & Greiner, 2007; Roy & Mccallum, 2001), variance reduction (Schein & Ungar, 2007), model change (Cai et al., 2014; Freytag et al., 2014; Käding et al., 2016; Settles & Craven, 2008), and min-max view active learning (Hoi et al., 2008; Huang et al., 2014). The principal difference among the above methods lies in how they measure the usefulness of unlabeled samples after retraining the model. For example, error reduce methods like EER (Roy & Mccallum, 2001) attempt to estimate the future generalization error as an indicator of the value of an instance while variance reduction approach (Schein & Ungar, 2007) turns to use the model variance as a measure of the informativeness. Similarly, model change algorithms seek various ways of defining such change, *e.g.* as gradient length (Settles & Craven, 2008), and choose the instance which leads to maximum change. The min-max view active learning directly measures the value of objective function during retraining procedure and selects the instance with minimum score in the worst case scenario. Recently, Yang & Loog (2016) proposed to improve the retraining-based algorithms by integrating the uncertainty information in the selection criterion.

We finally note that there exist close relationships between the proposed method and various active learning techniques, such as query-by-committee (QBC) (Seung et al., 1992), and variance minimization (Ji & Han, 2012; Ma et al., 2013; Yu et al., 2006). Their connections will be particularly explained in Subsection 3.5.

## 3. Maximizing Variance for Active Learning

We give a detailed description of the proposed method. We provide the full algorithm and introduce what is at the core of our method: so-called retraining information matrices (RIMs). Based on these RIMs, we introduce the two main types of variance and describe how these are fused into a single criterion for instance selection. In all of this, we focus on probabilistic classifiers. In Subsection 3.4, we show one way to adapt our method to a non-probabilistic classifier that does not directly provide a posterior probability estimate. We particularly focus on the SVM, which is the classifiers we are going to experiment with next to logistic regression. In Subsection 3.5, we analysis the connections of the proposed method and several existing active learning approaches. First however, we spend a few words on the specific active learning setting we consider.

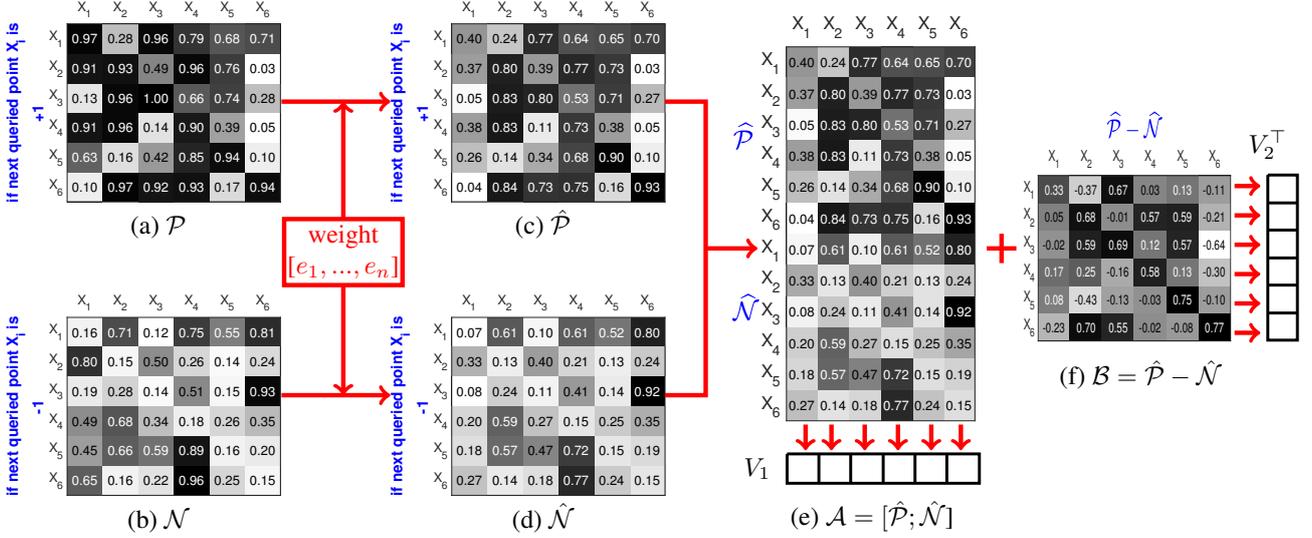

*Figure 1.* An overview of the proposed method MVAL. (a) retraining information matrix $\mathcal{P}$ represents that each of the next queried instance $x_i \in \mathcal{U}$ is labeled $+1$; (b) $\mathcal{N}$ means that each of the next queried instance $x_i \in \mathcal{U}$ is labeled $-1$; (c) and (d) are the weighted retraining information matrices of $\mathcal{P}$ and $\mathcal{N}$, respectively, where $[e_1, ..., e_n]$ are the defined weights; (e) and (f) correspond to two matrices $\mathcal{A}$ and $\mathcal{B}$, which are the combinations of $\hat{\mathcal{P}}$ and $\hat{\mathcal{N}}$. $V_1$ is the variance of each column in $\mathcal{A}$ while $V_2$ corresponds to the variance of each row in $\mathcal{B}$. $V_2^\top$ is the transpose of $V_2$. MVAL fuses $V_1$ and $V_2$ to evaluate the usefulness of unlabeled data.

## 3.1. Specific Setting

We study pool-based active learning in which the selection of individual instances to be labeled is sequential and myopic. This means that we assume we already have a large pool of unlabeled data with a small number of labeled data, and a single sample is selected for labeling at a time (Settles, 2010). We start with the binary classification problem, then present how to extend the proposed method to multi-class tasks in the following section, Section 4. We take $\mathcal{U}$ to be the pool of $n$ unlabeled instances $\{x_i\}_{i=1}^n$ and $\mathcal{L}$ to be the already labeled training set, where $y_i = \{+1, -1\}$ is the class label of $x_i$. $P_\mathcal{L}(y|x)$ represents the conditional probability of $y$ given $x$ on the basis of a classifier trained on $\mathcal{L}$.

## 3.2. Retraining Information Matrices

Figure 1 gives a pictorial overview of the proposed method. The proposed method can be used with different types of classifiers. In addition, Algorithm 1 summarizes the overall training procedure of MVAL for probabilistic classifiers. The proposed method generates two matrices $\mathcal{P}, \mathcal{N}$, with the purpose of recording the probability of all unlabeled instances after each retraining procedure. We first assume that the next queried instance is labeled as $+1$, we then extend the current labeled set $\mathcal{L}^+ = \mathcal{L} \cup \{x_i, +1\}$, retrain the classifier on $\mathcal{L}^+$, and calculate the conditional probability $P_{\mathcal{L}^+}(+1|x_j)$ for all $x_j \in \mathcal{U}$. Each $x_i \in \mathcal{U}$

---

**Algorithm 1** Maximizing Variance for Active Learning

1: **Input:** Labeled data $\mathcal{L}$, unlabeled data $\mathcal{U}$
2: **repeat**
3:  Train on $\mathcal{L}$ and calculate entropy $e_j$ for all $x_j \in \mathcal{U}$;
4:  For each $x_i \in \mathcal{U}$, retrain on $\mathcal{L}^+ = \mathcal{L} \cup \{x_i, +1\}$, let $\mathcal{P}_{i,j} = P_{\mathcal{L}^+}(+1|x_j)$, $x_j \in \mathcal{U}$;
5:  For each $x_i \in \mathcal{U}$, retrain on $\mathcal{L}^+ = \mathcal{L} \cup \{x_i, -1\}$, let $\mathcal{N}_{i,j} = P_{\mathcal{L}^+}(+1|x_j)$, $x_j \in \mathcal{U}$;
6:  Obtain weighted $\hat{\mathcal{P}}$ and $\hat{\mathcal{N}}$ and compute the variance using Eq. 3;
7:  Query the instance $x^*$ with maximum variance and label it $y^*$, update $\mathcal{L} \leftarrow \mathcal{L} \cup \{x^*, y^*\}, \mathcal{U} \leftarrow \mathcal{U} \backslash \{x^*\}$;
8: **until** Stopping criterion is satisfied

---

is used to retrain the model, resulting in a matrix $\mathcal{P}$ of size $n \times n$, where each element $(i, j)$ in $\mathcal{P}$ is assigned $P_{\mathcal{L}^+}(+1|x_j)$. For example, assuming that $\mathcal{U}$ consists of six unlabeled samples $x_i, i = 1, 2, \ldots, 6$, we could get the matrix $\mathcal{P}$ in Figure 1a. Equivalently, if we categorize all of the next queried instances as $-1$, we retrain the model with $\mathcal{L}^+ = \mathcal{L} \cup \{x_i, -1\}$ for all $x_i \in \mathcal{U}$, we can construct a matrix $\mathcal{N}$ that contains the elements $\mathcal{N}_{i,j} = P_{\mathcal{L}^+}(+1|x_j)$, of which an example is shown in Figure 1b.

The matrices $\mathcal{P}$ and $\mathcal{N}$ are the RIMs that collect and preserve the output information over the unlabeled pool during the retraining process. We note here already that since we assign all the elements in the RIMs the value of $P_{\mathcal{L}^+}(+1|x_j)$, for the variance computation it will make no difference if we change the value to $P_{\mathcal{L}^+}(-1|x_j)$ since we are dealing with binary classification problem and $P_{\mathcal{L}^+}(-1|x_j) = 1 - P_{\mathcal{L}^+}(-1|x_j)$.

We subsequently introduce a entropy weighted version of these RIMs, similar to the correction strategy that was presented in (Yang & Loog, 2016), which reflects the ideas behind uncertainty sampling (Lewis & Gale, 1994), in which the selection mechanism is purely based on the current classifier trained on the original $\mathcal{L}$ (rather than retrained on $\mathcal{L}^+$). With this weighting we aim to trade off uncertainty due to instability of an instance and uncertainty due to closeness to the decision boundary. Specifically, we firstly compute the pre-retraining entropy $e_j = -\sum_{y_j \in \pm 1} P_{\mathcal{L}}(y_j|x_j) \log(P_{\mathcal{L}}(y_j|x_j)), j = 1, \ldots, n$ and subsequently obtain two weighted matrices $\hat{\mathcal{P}}$ and $\hat{\mathcal{N}}$, where $\hat{\mathcal{P}}_{i,j} = e_j \times \mathcal{P}_{i,j}$ and $\hat{\mathcal{N}}_{i,j} = e_j \times \mathcal{N}_{i,j}$.

### 3.3. Variance Computations

The two information matrices we compute do not lead directly to a selection criterion that we can determine for each instance. Here it is where we consider particular variances derived from these RIMs. As shown in Figure 1, we firstly construct two different matrices by combining $\hat{\mathcal{P}}$ and $\hat{\mathcal{N}}$. The first one concatenates $\hat{\mathcal{P}}$ and $\hat{\mathcal{N}}$ column-wise, resulting in a new matrix $\mathcal{A} = [\hat{\mathcal{P}}; \hat{\mathcal{N}}]$ of size $2n \times n$ in Figure 1e. We obtain second matrix $\mathcal{B} = \hat{\mathcal{P}} - \hat{\mathcal{N}}$ of size $n \times n$ by subtracting $\hat{\mathcal{N}}$ from $\hat{\mathcal{P}}$, as illustrated in Figure 1f.

For matrix $\mathcal{A}$, the column-wise variance is derived to form a vector denoted as $V_1$, in which the $j$-th element corresponding to the variance in the $j$-th column is calculated by

$$V_{1,j} = \frac{1}{2n-1} \sum_{i=1}^{2n} (\mathcal{A}_{i,j} - \frac{1}{n} \sum_{i=1}^{2n} \mathcal{A}_{i,j})^2, \quad j = 1, \ldots, n \tag{1}$$

where $\mathcal{A}_{i,j}$ represents the value of element $(i, j)$ in $\mathcal{A}$. In contrast, we compute the row-wise variance for matrix $\mathcal{B}$, which is stored in the vector $V_2$, i.e., the variance in the $i$-th row is calculated by

$$V_{2,i} = \frac{1}{n-1} \sum_{j=1}^{n} (\mathcal{B}_{i,j} - \frac{1}{n} \sum_{j=1}^{n} \mathcal{B}_{i,j})^2, \quad i = 1, \ldots, n \tag{2}$$

Herein $\mathcal{B}_{i,j}$ is the value of element $(i, j)$ of $\mathcal{B}$.

The reasons for creating the matrices $\mathcal{A}$ and $\mathcal{B}$ and the way of calculating their variances $V_1$ and $V_2$ are the following. Firstly, the variance of each column of $\mathcal{A}$ is important since it captures the variations of unlabeled samples when we query a different sample or label it a different category. Each column of $\hat{\mathcal{P}}$ and $\hat{\mathcal{N}}$ represents the scenario that we choose different instances as the next candidate. Concatenating $\hat{\mathcal{P}}$ and $\hat{\mathcal{N}}$ column-wise like $\mathcal{A}$ indicates that we attach totally contradictory label to the next queried sample. Therefore, $V_1$, which represents the instability or uncertainty when the next queried sample or its corresponding label changes, is a measure of the informativeness. Secondly, the element $(i, j)$ in $\mathcal{B}$ represents the difference of $P_{\mathcal{L}}(+1|x_j)$ caused by assigning $x_i$ a totally different label. If $x_i$ is representative of $x_j$, e.g., $x_i$ and $x_j$ come close to each other or belong to the same cluster, element $(i, j)$ in $\mathcal{P}$ should vary markedly from $(i, j)$ in $\hat{\mathcal{N}}$ since $x_i$ is labeled differently and the element $(i, j)$ of $\mathcal{B}$ should significantly differ from zero. Hence, the variance of the row of $\mathcal{B}$ indicates the impact of an instance over other unlabeled data when its annotated label varies. $V_2$ can be seen as a measure of the representativeness. Finally, since the variances are calculated over weighted $\hat{\mathcal{P}}$ and $\hat{\mathcal{N}}$, both $V_1$ and $V_2$ essentially take advantage of the uncertainty information provided by the entropy.

Now we need to fuse $V_1$ and $V_2$ to sort the unlabeled data. In this paper, we use a simple approach: element-wise multiplication $V_1 \cdot V_2$. We propose the maximizer of this product as our new selection criterion for active learning:

$$\arg \max_{x \in \mathcal{U}} V_1 \cdot V_2 \tag{3}$$

Since $V_1$ and $V_2$ can measure the informativeness and representativeness, respectively, MVAL is able to select the samples which are both informative and representative.

### 3.4. Adaptation to SVM

For classifiers which do not produce a probabilistic output, we can adapt the proposed method by using their decision values. The particular example we focus on, which will also be used in our experiments, is the SVM. Directly using the decision value $f(x_j)$ as the element of the RIMs leads the variance estimates to be overly sensitive to decision values which may be extremely large or small and empirical experiments indeed show poor results for the above choice. Therefore, similar to the scaling in (Platt et al., 1999), we are going to transform the decision values into a type of probabilistic outputs. We do not directly rely on Platt scaling, however, because the limited amount of labeled training data, especially in the beginning of active learning, fails to produce stable estimates for these probabilities. Instead, we take a fixed sigmoidal transfer function $(1 + \exp(-f(x)))^{-1}$ to transform decision values into probabilities. This sigmoidal transfer corresponds to the probabilistic output one would obtain if instead of the hinge loss, one would plugs in a logistic loss function that respects the same margin as the original hinge loss.

In order to obtain weighted RIM, $\hat{\mathcal{P}}$ and $\hat{\mathcal{N}}$, we also need estimate the weight. Instead of firstly transforming the probability and then computing the entropy, we adopt the $e_j = \exp(-|f(x_j)|)$ as the weight, which means that the instance that is nearest to the decision margin receives the highest weight. The proposed method can be easily adapted to other classifiers. In the experimental section, we validate the performance of the proposed method with logistic regression and SVM, respectively.

### 3.5. Comparisons and Connections

The proposed method mainly preserves the relevant information during the retraining procedure and creates RIMs to capture the variance of unlabeled samples. Indeed, there exist several connections to other active learning approaches, such as QBC (Mamitsuka, 1998; McCallumzy & Nigamy, 1998; Seung et al., 1992), bootstrap-local variance method (BSLV) (Saar-Tsechansky & Provost, 2004), and variance-minimization approaches (Ji & Han, 2012; Ma et al., 2013; Schein & Ungar, 2007; Yu et al., 2006).

MVAL shares a similar idea with QBC but performs slightly different. QBC approaches first constitute a committee of models and then measures the disagreement among the different committee members. Similarly, MVAL can be seen as a specific version of QBC since it also makes use of a number of committee (such as the model re-trained on $\mathcal{L}^+ = \mathcal{L} \cup \{x_i, \pm 1\}$ and estimates the variance as the disagreement. The slight differences lies in: (1) typical QBC algorithms use Gibbs algorithm (Seung et al., 1992) or re-sampling method such as boosting and bagging (Mamitsuka, 1998) to generate the a committee, while MVAL directly utilizes the current training data and one more unlabeled sample with its potential labels. The presence of additional unlabeled samples make the committee more flexible, which can increase the levels of disagreements among committees; (2) QBC normally employs vote entropy or KL divergence (McCallumzy & Nigamy, 1998) to measure the disagreement, whereas MVAL designs two particular variances based on RIMs as the disagreement. And these variances correspond to the informativeness and representativeness, respectively. Therefore, one advantage of MVAL over QBC is that QBC is not able to estimate the representativeness of samples.

MVAL is also different from the BSLV (Saar-Tsechansky & Provost, 2004), which bootstraps from the already labeled data and calculates the variance of each unlabeled instance. Several differences exist: (1) BSLV uses bootstrap sampling to generate various models; (2) BSLV only calculate a kind of variance which is slightly similar to the $V_1$ in MVAL; (3) BSLV is not a deterministic selection algorithm since it normalizes the variance as a randomly selection distribution.

There is a major difference between MVAL and several variance-minimization methods such as transductive experimental design (TED) (Yu et al., 2006), variance reduc-tion (Schein & Ungar, 2007; Zhang & Oles, 2000) and graph-based variance minimization (Ji & Han, 2012; Ma et al., 2013). The sharpest distinction is that MVAL prefers the instance whose individual variance is the largest while these variance-minimization algorithms favour the sample which leads to a minimum variance of a statistical model. For example, experimental design approaches aim to minimize the output variance of some specific statistical models to sequentially reduce the future generalization error. Graph-based methods in (Ji & Han, 2012; Ma et al., 2013) focus on the tasks where the graph structure is available without the feature representation. Based on the Gaussian random field classifier, it selects the nodes which minimizes expected prediction variance once labeled. Expected variance reduction (EVR) (Schein & Ungar, 2007), which also belongs to the retraining-based active learning, obtains an approximation of the model variance during the retraining process. Unlike experimental design, EVR and graph-based algorithms, MVAL directly estimates the variance of each unlabeled sample introduced by retraining with different training data instead of calculating the model variance. Another dissimilarity is that TED (Yu et al., 2006) and two graph-based algorithms (Ji & Han, 2012; Ma et al., 2013) do not make use of the label information of the queried samples. This means that these methods can not benefit from the feedback information which comes from the human annotator. On the contrary, our method utilizes the label information to update the model in each iteration. As shown in (Zhen & Yeung, 2010), the label information can provide useful hints for active learning. Therefore, these methods in (Ji & Han, 2012; Ma et al., 2013; Yu et al., 2006) is less competitive than the proposed method. We will verify this through empirical experiments in Subsection 5.3 (See Table 3).

## 4. MVAL for Multi-class Classification

In this section, we extend MVAL to multi-class classification problems. A simple approach to addressing this issue is to reduce a multi-class task as multiple binary subtasks using one-vs-all strategy. As Yang et al. (2015) addressed, however, this may lead to a degradation of the performance of active learners since it is difficult to fuse the results across multiple binary classifiers. We present an alternative approach, which also follows the retraining procedures and keeps record of relevant information. When it comes to the multiclass case, the main challenges are how to generate the RIMs and how to construct the variances.

For binary problem, RIMs are 2D matrices since each element of RIM is a single value $P_{\mathcal{L}^+}(+1|x_j)$. Nevertheless, for a multiclass task of $K$ classes $\{1, 2, \ldots, K\}$, we need record all the posterior probabilities $P_{\mathcal{L}^+}(l|x_j)$, $l = 1, 2, \ldots, K$ when the model is retrained. The advan-tage is that no information is discarded during the retraining process. Thus, we can constitute $K$ different RIMs of size $n \times n \times K$, where $n$ is the number of unlabeled samples and the third references dimension corresponds to the posterior probabilities. These RIMs, which are 3D matrices, are denoted as $R_k$, $k = 1, 2, \ldots, K$. The whole procedure is shown in Figure 2. First, the model is retrained by adding each unlabeled instance with pseudo label $k$, resulting in a RIM $R_k$. The element $(i, j, l)$ of $R_k$ is the posterior probability $P_{\mathcal{L} \cup \{x_i, k\}}(l|x_j)$, $k, l = 1, 2, \ldots, K$, $i, j = 1, 2, \ldots, n$. Next, similar to the weighting scheme used in binary case, each $R_k$ is transformed to weighted $\hat{R}_k$. The weighting function we use here is the margin sampling (Settles, 2010), which is equivalent to entropy-based uncertainty sampling in binary case but demonstrates much better performance than entropy-based uncertainty sampling on multi-class tasks (Joshi et al., 2009). Note that the original criterion of margin sampling is finding the minima of $\min_{x_j} (P_{\mathcal{L}}(\hat{y}_1|x_j) - P_{\mathcal{L}}(\hat{y}_2|x_j))$, where $\hat{y}_1$ and $\hat{y}_2$ are two class labels which rank first and second, respectively based on the model trained on current labeled data $\mathcal{L}$. To be consistent to our objective of querying the samples with maximum variance, we use $exp(-(P_{\mathcal{L}}(\hat{y}_1|x_j) - P_{\mathcal{L}}(\hat{y}_2|x_j)))$ as the weight $e_j$ of sample $x_j$. More precisely, weighted $\hat{R}_k$ are obtained as $\hat{R}_k(i, j, l) = e_j \times R_k(i, j, l)$.

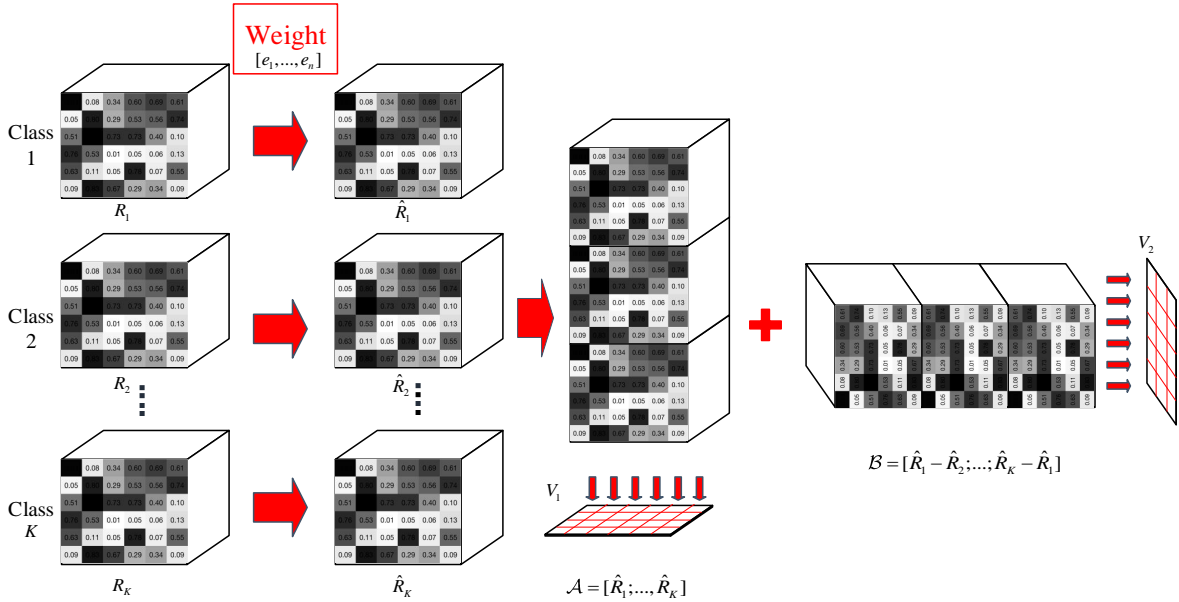

*Figure 2.* An illustration of MVAL for multi-calss classification. $R_k$ and $\hat{R}_k$ are the original and weighted 3D retraining information matrices, respectively, where $k = 1, 2, \ldots, K$ and $[e_1, \ldots, e_n]$ are the predefined weights of unlabeled samples. $\mathcal{A}$ and $\mathcal{B}$ are two combinations of $\hat{R}_k$ on which two kinds of variance $V_1$ and $V_2$ are computed. MVAL fuses $V_1$ and $V_2$ to evaluate the usefulness of unlabeled data.

Finally, we estimate the variance of each unlabeled sample on the basis of these 3D RIMs. As shown in Fig 2,

two new matrices are constructed as follows: (1) all the weighted $\hat{R}_k$ are concatenated column-wisely to form a matrix $\mathcal{A} = [\hat{R}_1; \hat{R}_2; \ldots; \hat{R}_K]$ of size $nK \times n \times K$; (2) in the binary case, we measure the difference between two RIMs $\hat{\mathcal{P}}$ and $\hat{\mathcal{N}}$ to estimate the representativeness. However, we own $K$ different RIMs instead of two RIMs in the multi-class case. Here we propose to evaluate the differences between all adjacent pairs $\hat{R}_k$ and $\hat{R}_{k+1}$ and concatenate these results row-wisely, resulting in a new matrix $\mathcal{B} = [\hat{R}_1 - \hat{R}_2, \hat{R}_2 - \hat{R}_3, \ldots, \hat{R}_K - \hat{R}_1]$ of size $n \times nK \times K$. An alternative approach, which considers all the paired difference among $\hat{R}_i$ and $\hat{R}_j$, $i, j = 1, 2, \ldots, K$, has a heavy computational cost, especially when $K$ is large. Therefore, we only consider the difference of adjacent $\hat{R}_k$ and $\hat{R}_{k+1}$.

Even though a different ordering of the classes will, in principle, lead to a potentially different outcome, preliminary experimental results show that the ordering has a negligible effect on the overall performance of the proposed method.

Similarly to the binary case, the column-wise variance of $\mathcal{A}$ and the row-wise variance of $\mathcal{B}$ are calculated. Note that the $\mathcal{A}$ and $\mathcal{B}$ are 3D matrices, which means that the variances of $\mathcal{A}$ and $\mathcal{B}$ are still 2D matrices. The idea used here is that we first calculate the column-wise variance of $\mathcal{A}$ according to the first dimension and then measure its mean over the third dimension as $V_1$. On the other hand, $V_2$ are firstly computed on the second dimension and then averaged on the third dimension. In the end, the final selection crite-ria of multiclass MVAL is the same with Equation 3: the element-wise multiplication of $V_1$ and $V_2$. Accordingly, $V_1$ and $V_2$ indicate the informativeness and representativeness, respectively.

# 5. Experiments with Binary Classification

We empirically compare our proposed method with state-of-the-art active learning algorithms. Extensive results on 45 binary benchmark datasets demonstrate the effectiveness and robustness of our method. We start by a brief description of the various test datasets. Subsequently, we examine how the proposed method works in comparison with other active learning methods using logistic regression and SVM, respectively.

## 5.1. Datasets

To evaluate the performance of different active learning algorithms, 45 benchmark datasets are used as the test bed. Some basic information about the datasets after preprocessing is shown in Table 1. Many of these datasets are commonly used in other active learning experiments, such as the 20 Newsgroups dataset in (Yu et al., 2006; Zhu et al., 2003) and the Letter Recognition dataset in (Huang et al., 2014). A large number of datasets originally comes from the UCI Machine Learning Repository (Lichman, 2013)[1], such as australian, mammographic, vehicle, wdbc and so on. Here, however, we use the preprocessed version such as presented in (Fernández-Delgado et al., 2014). Datasets containing more than two classes are converted to binary datasets. Specifically, six pairs of letters from Letter Recognition dataset, i.e., DvsP, EvsF, IvsJ, MvsN, VvsY and UvsV, are constructed as the binary datasets. Seven binary datasets are taken from the 20 Newsgroups dataset (Lang, 1995), which is a commonly used collection for text classification[2]. The first three datasets, baseball vs. hockey, pc vs. mac, and misc vs. atheism, are also used for comparison in (Zhu et al., 2003). The remaining four datasets, autos, motorcycles, baseball, hockey, are pre-processed according to (Yu et al., 2006)[3]. Since a one-against-all scheme is used to create the above four binary datasets, it represents a case of fairly imbalanced binary classification problems. The MNIST database (LeCun et al., 1998)[4] is a commonly used handwritten digit dataset and we construct three paired datasets based on it, i.e., 3vs5, 5vs8 and 7vs9, to test the performance of the different active learners.

For computational efficiency, we apply random sub-sampling and principal component analysis (PCA) on some datasets to reduce both the number of data points and the size of feature dimensionality.

## 5.2. Data Split and Initial Labeled Set

We randomly split each dataset into training and test set of equal size. We consider a difficult setting for active learning and start out with only two labeled instances at the very beginning. We randomly labeled one example of the positive class and one example of the negative class from the training set. For each active learning algorithm, the experiment is repeated 10 times on each real-world dataset, followed by a report of the average performance. Active learning is terminated when 100 samples are labeled on all of the datasets, except on those datasets that have too few instances to leave a properly sized test set.

## 5.3. Results using Logistic Regression

Many active learning algorithms are derived using particular classifiers. For example, the simple margin (Tong & Koller, 2002) approach was created based on SVM, while QUIRE (Huang et al., 2014) was developed using ridge regression. In evaluating our active learning method, we benchmark against methods that either have been designed for the same classifiers or can be easily adapted to the same classifiers. In this subsection, we firstly conduct experiments using active learning algorithms whose base classifier is logistic regression. Subsection 5.4 then presents experiments with active learning methods that rely on SVMs.

The following state-of-the-art active learning algorithms based on logistic regression are considered in addition to the standard baseline, i.e., random sampling (RS for short).

- **BSLV**: Bootstrap-LV algorithm, which bootstraps from the labeled data and estimates the variance as the randomly sampling probability distribution (Saar-Tschansky & Provost, 2004);

- **US**: an uncertainty sampling approach, which queries the example with highest entropy (Lewis & Gale, 1994; Settles, 2010);

- **EER**: Expected Error Reduction, which selects the sample with minimum future generalization error (Roy & Mccallum, 2001);

- **UEER**: Uncertainty based EER, an improved version of EER using the uncertainty information (Yang & Loog, 2016);

- **MLI**: Minimum Loss Increase, which switches from the square loss of QUIRE (Huang et al., 2014) to the logistic loss.

- **BMDR**: Batch mode active learning, which queries

---

*Table 1.* Datasets information after pre-processing: the number of instances (# Ins) and the feature dimensionality (# Fea)

| Dataset (# Ins, # Fea) | Dataset (# Ins, # Fea) | Dataset (# Ins, # Fea) |
|---|---|---|
| fertility (100, 9) | wdbc (569, 31) | 3vs5 (1500, 784) |
| ac-inflam (120, 6) | hill (606, 100) | 5vs8 (1500, 784) |
| acute (120, 6) | breast (683, 10) | 7vs9 (1500, 784) |
| wine (178, 13) | australian (690, 14) | IvsJ (1502, 16) |
| parkinsons (195, 22) | wisc (699, 9) | EvsF (1543, 16) |
| sonar (208, 60) | blood (748, 4) | UvsV (1550, 16) |
| glass (214, 9) | diabetes (768, 8) | MvsN (1575, 16) |
| hepatitis (155, 19) | pima (768, 8) | VvsY (1577, 16) |
| heart (270, 13) | ooctis2f (912, 25) | DvsP (1608, 16) |
| vc2 (310, 6) | tictactoe (958, 9) | pc-mac (1945, 500) |
| liver (345, 6) | mammographic (961, 5) | base-hockey (1993, 500) |
| ionosphere (351, 34) | mushrooms (1000, 112) | autos (3970, 8014) |
| vehicle (435, 18) | ozone (1000, 72) | motorcycles (3970, 8014) |
| musk1 (476, 166) | splice (1000, 60) | baseball (3970, 8014) |
| cylinder (512, 35) | misc-atheism (1427, 500) | hockey (3970, 8014) |

discriminative and representative samples. The batch size is set as 1 in this comparison (Wang & Ye, 2013).

We use the $L_2$ regularized logistic regression method implemented in the LIBLINEAR package (Fan et al., 2008) as the classification model for all the algorithms that we compare. Default parameters are used and the penalty parameter $C$ is set to 100 in all the experiments. For BMDR, a trade-off parameter $\beta$ is used to balance the informativeness and representativeness. We carefully tuned this parameter and set $\beta = 1$ which shows the best average performance over all the datasets. We consider learning curves, which plot the classifier accuracy on test data as a function of the number of labeled training examples. The area under the learning curve (ALC) is then used as the performance measure (Cook & Krishnan, 2015).

The performance of seven active learning approaches based on logistic regression on our 45 datasets are presented in Table 2. A paired $t$-test at a 95% significance level is adopted to evaluate whether two methods are significantly different from each other. For each dataset, the active learning methods which perform the best or are able to compete with the best one are highlighted in bold face and coloured. Some criteria, like average ALC and average ranking, are also reported in Table 2. The win/tie/loss counts are also provided based on paired $t$-test at 95% significance level. All the datasets are sorted in ascending order based on the average ALC scores of random sampling, or in other words, they are sorted from difficult to easy classification tasks from the perspective of logistic regression.

We see that the proposed method achieves the best performance in terms of average ALC and average ranking. MVAL obtains the highest average ALC score 0.839 while the second best one is 0.828 achieved by UEER. The aver-age ranking of MVAL is smaller than 2 and, in most cases, MVAL ranks in the first or second position. There are 34 datasets on which MVAL obtains the best performance or one not significantly different from the best scoring other method. The second one is MLI in 12 datasets. Generally, MVAL demonstrates highly competitive performance in comparison with other methods over all the datasets, *e.g.* the win/tie/loss counts of MVAL versus the second best one UEER is 29/14/2. And this value of MVAL versus US is 36/8/1. This confirms the effectiveness of the proposed method. We also observe that though US is a quite simplistic approach, it still outperforms some sophisticated methods like BMDR and MLI with regards to average ALC. There are some datasets on which many active learning methods actually lose when compared to random sampling. For example, random sampling outperforms MLI on 11 datasets. It therefore is very interesting to note that MVAL never performs worse than random sampling over all 45 sets and only reaches a tie on 1 of the datasets.

Figure 3a presents the average accuracy of the first 30 labeled instances over all the 45 datasets for logistic-based active learning algorithms in. MVAL clearly outperforms other methods, while UEER is a good second, being slightly better than EER and US.

To further investigate the distinction between our variance-maximization method and variance-minimization methods, we also construct experiments to empirically compare their performance. Random sampling (RS) and two graph-based methods $V$-optimality ($V$-opt) (Ji & Han, 2012) and $\Sigma$-optimality ($\Sigma$-opt) (Ma et al., 2013) are included in this comparison, followed by two experimental design algorithms, TED (Yu et al., 2006) and Logistic Bound (Gu et al., 2014). As shown in Table 3, we only report the average performances of compared methods. A detailed description of the performances on each single dataset is included in the appendix. Note that we only show the results on 41 binary datasets since there are four relatively large datasets, i.e. autos, motorcycles, baseball and hockey, on which we can not manage to conduct $V$-opt and $\Sigma$-opt due to high computational cost involved in computing the inverse matrix. Still, we find that the proposed method obtains the best average performance. Logistic Bound also has a very competitive performance and is far better than TED. This is because that (1) Logistic Bound can be seen as a weighted version of TED where the weights are closely related to the entropy of unlabeled samples and (2) Logistic Bound takes into account uncertainty information derived form label information while TED does not utilize this kind of label information. However, our method still outperforms Logistic Bound on 21 datasets and only fails on 7 datasets.

*Table 2.* Performance comparisons of active learning algorithms in terms of the areas under the learning curve (ALC), based on logistic regression. "Average ALC" reports the average ALC scores over all the datasets. "Average Ranking" shows the average ranking within the compared methods. "Win Times" is the number of datasets on which an algorithm achieving the best or comparable performance. "W/T/L MVAL VS" represents the win/tie/loss counts of MVAL versus other algorithms over all the datasets. Similarly, "W/T/L VS RS" shows the win/tie/loss counts of other methods versus random sampling.

| | RS | BSLV | US | EER | UEER | MLI | BMDR | MVAL |
|---|---|---|---|---|---|---|---|---|
| hill | 0.583 | 0.599 | 0.591 | **0.619** | 0.592 | 0.616 | **0.622** | 0.621 |
| cylinder | 0.596 | 0.585 | 0.597 | **0.617** | 0.601 | **0.616** | 0.587 | 0.602 |
| liver | **0.628** | 0.612 | 0.581 | **0.629** | 0.606 | 0.600 | 0.621 | 0.621 |
| splice | 0.651 | 0.663 | **0.672** | **0.672** | 0.668 | 0.644 | 0.646 | **0.671** |
| occtris2f | 0.686 | 0.681 | 0.671 | 0.688 | 0.679 | 0.685 | 0.666 | **0.699** |
| musk1 | 0.688 | 0.690 | 0.699 | 0.704 | **0.714** | 0.703 | 0.704 | **0.713** |
| sonar | 0.688 | 0.701 | 0.698 | **0.707** | 0.691 | 0.690 | 0.696 | **0.704** |
| pcmac | 0.688 | 0.710 | 0.693 | 0.686 | 0.677 | 0.719 | **0.722** | 0.724 |
| religionathesim | 0.689 | 0.701 | 0.709 | 0.679 | 0.686 | 0.643 | **0.714** | 0.698 |
| pima | 0.704 | 0.717 | 0.714 | 0.700 | 0.708 | 0.699 | 0.685 | **0.725** |
| fertility | 0.707 | **0.729** | 0.705 | 0.720 | 0.719 | **0.728** | 0.665 | **0.728** |
| diabetes | 0.708 | 0.722 | 0.718 | 0.727 | 0.726 | 0.728 | 0.711 | **0.731** |
| blood | 0.727 | **0.738** | 0.721 | **0.740** | 0.705 | 0.734 | 0.661 | **0.739** |
| hepatitis | 0.727 | 0.732 | **0.773** | **0.768** | 0.758 | 0.742 | 0.760 | **0.767** |
| heart | 0.758 | 0.789 | 0.783 | 0.782 | 0.782 | **0.796** | 0.786 | 0.791 |
| baseball | 0.759 | 0.793 | 0.850 | 0.765 | **0.871** | 0.836 | 0.781 | 0.857 |
| autos | 0.760 | 0.793 | 0.845 | 0.769 | **0.871** | 0.839 | 0.779 | **0.866** |
| motorcycles | 0.765 | 0.798 | 0.858 | 0.777 | 0.883 | 0.853 | 0.796 | **0.888** |
| basehockey | 0.766 | 0.784 | 0.780 | 0.736 | 0.749 | 0.770 | 0.816 | **0.822** |
| hockey | 0.783 | 0.823 | 0.886 | 0.786 | 0.899 | 0.872 | 0.811 | **0.911** |
| mammographic | 0.783 | 0.791 | 0.770 | 0.772 | **0.801** | 0.776 | **0.796** | 0.793 |
| australian | 0.785 | 0.832 | 0.839 | 0.818 | 0.832 | 0.839 | 0.837 | **0.848** |
| ionosphere | 0.797 | 0.801 | 0.769 | **0.823** | 0.800 | 0.666 | 0.790 | **0.822** |
| parkinsons | 0.811 | 0.819 | 0.824 | 0.816 | 0.818 | **0.828** | 0.816 | 0.825 |
| vc2 | 0.812 | 0.821 | 0.813 | 0.811 | 0.813 | **0.826** | 0.794 | 0.814 |
| letterIJ | 0.849 | 0.859 | 0.861 | 0.874 | 0.824 | 0.851 | 0.878 | **0.891** |
| 5vs8 | 0.855 | 0.877 | 0.894 | 0.907 | 0.906 | 0.846 | 0.898 | **0.914** |
| 7vs9 | 0.856 | 0.891 | 0.901 | **0.918** | **0.916** | 0.849 | 0.909 | **0.919** |
| vehicle | 0.858 | 0.870 | 0.881 | 0.871 | 0.886 | 0.866 | 0.877 | **0.900** |
| letterVY | 0.864 | 0.856 | 0.881 | 0.880 | 0.881 | 0.860 | 0.880 | **0.893** |
| 3vs5 | 0.866 | 0.871 | 0.886 | **0.906** | 0.898 | 0.860 | 0.883 | 0.902 |
| ozone | 0.877 | 0.875 | 0.883 | 0.833 | **0.889** | 0.872 | 0.877 | **0.887** |
| tictactoe | 0.896 | 0.893 | 0.898 | **0.907** | **0.905** | 0.849 | 0.875 | 0.905 |
| wine | 0.899 | 0.925 | 0.923 | 0.938 | 0.942 | 0.936 | 0.934 | **0.948** |
| glass | 0.899 | 0.908 | 0.904 | 0.907 | 0.912 | **0.915** | 0.908 | 0.913 |
| letterMN | 0.911 | 0.925 | **0.941** | 0.941 | **0.945** | 0.928 | 0.934 | 0.941 |
| wdbc | 0.916 | 0.952 | 0.952 | 0.951 | 0.953 | **0.956** | 0.942 | 0.954 |
| mushrooms | 0.931 | 0.953 | 0.973 | 0.969 | 0.972 | 0.971 | 0.957 | 0.976 |
| letterEF | 0.934 | 0.948 | 0.958 | 0.954 | 0.960 | 0.956 | 0.953 | **0.962** |
| letterDP | 0.938 | 0.949 | 0.961 | 0.962 | 0.967 | 0.966 | 0.952 | **0.970** |
| breast | 0.943 | 0.960 | 0.960 | 0.957 | 0.962 | **0.964** | 0.950 | 0.961 |
| ac-inflam | 0.947 | 0.972 | **0.984** | 0.979 | **0.983** | 0.979 | 0.966 | **0.981** |
| wisc | 0.949 | 0.951 | 0.954 | 0.952 | 0.956 | **0.956** | 0.947 | 0.956 |
| letterUV | 0.949 | 0.962 | 0.969 | 0.972 | **0.977** | **0.975** | 0.964 | **0.978** |
| acute | 0.967 | 0.975 | **0.991** | 0.955 | 0.978 | **0.993** | 0.984 | **0.991** |
| Average ALC | 0.803 | 0.818 | 0.825 | 0.819 | 0.828 | 0.818 | 0.816 | **0.839** |
| Average Ranking | 6.84 | 5.11 | 4.31 | 4.49 | 3.71 | 4.42 | 5.29 | **1.82** |
| Win Times | 1 | 2 | 5 | 11 | 10 | 12 | 4 | **34** |
| W/T/L MVAL VS | 44/1/0 | 37/6/2 | 36/8/1 | 31/12/2 | 29/14/2 | 32/6/7 | 41/3/1 | - |
| W/T/L VS RS | - | 37/4/4 | 36/4/5 | 35/3/7 | 36/2/7 | 30/4/11 | 33/2/10 | **44/1/0** |

*Table 3.* Performance comparisons of the proposed method versus variance-minimization algorithms on 41 binary datasets.

| | RS | $V$-opt | $\Sigma$-opt | TED | Logistic Bound | MVAL |
|---|---|---|---|---|---|---|
| Average ALC | 0.807 | 0.806 | 0.815 | 0.810 | 0.827 | **0.834** |
| Average Ranking | 4.93 | 4.05 | 3.63 | 4.49 | 2.29 | **1.61** |
| Win Times | 1 | 6 | 6 | 3 | 18 | **30** |
| W/T/L MVAL VS | 40/1/0 | 33/7/1 | 34/3/4 | 34/6/1 | 21/13/7 | - |
| W/T/L VS RS | - | 27/3/11 | 28/6/7 | 20/8/13 | 35/2/4 | **40/1/0** |

Let us, for completeness, also report the overall performance of each component of MVAL based on the original RIMs and the weighted RIMs. The average ALC values over all the datasets are provided in Table 4. $V_1$, $V_2$, and $V_1 \cdot V_2$ represent the different types of variance as introduced in Subsection 3.3. The fusion of $V_1$ and $V_1 \cdot V_2$ significantly outperforms each single term on both original RIMs and weighted RIMs based on a paired $t$-test at a 95% significance level, which demonstrates the advantage of combining the informativeness introduced by $V_1$ and the representativeness carried by $V_2$. We observe that the same kind of variance on weighted RIMs markedly exceeds that on original RIMs. For example, a paired $t$-test shows that the performance of $V_1 \cdot V_2$ on $(\hat{\mathcal{P}}, \hat{\mathcal{N}})$ surpasses that on $(\mathcal{P}, \mathcal{N})$ at a 95% significance level. It is also the same situation for $V_1$ and $V_2$. This demonstrates that our proposed weighting scheme is able to enhance the performance of active learners.

*Table 4.* Average ALC of components of MVAL over all the datasets using logistic regression. $V_1$, $V_2$, and $V_1 \cdot V_2$ represent the different types of variance. $(\mathcal{P}, \mathcal{N})$ and $(\hat{\mathcal{P}}, \hat{\mathcal{N}})$ represent the original RIMs and the weighted RIMs, respectively.

| variance RIMs | $V_1$ | $V_2$ | $V_1 \cdot V_2$ |
|---|---|---|---|
| $(\mathcal{P}, \mathcal{N})$ | 0.827 | 0.786 | 0.833 |
| $(\hat{\mathcal{P}}, \hat{\mathcal{N}})$ | 0.831 | 0.815 | 0.839 |

### 5.4. Results using SVM

Support vector machines are a popular classification method used in active learning (Hoi et al., 2008; Kremer et al., 2014; Tong & Koller, 2002). Here we compare our method with random sampling and several active learning approaches which are used in combination with SVM. These methods are named as follows:

- SIMPLE: simple margin, which selects the example closest to the decision boundary (Tong & Koller, 2002);

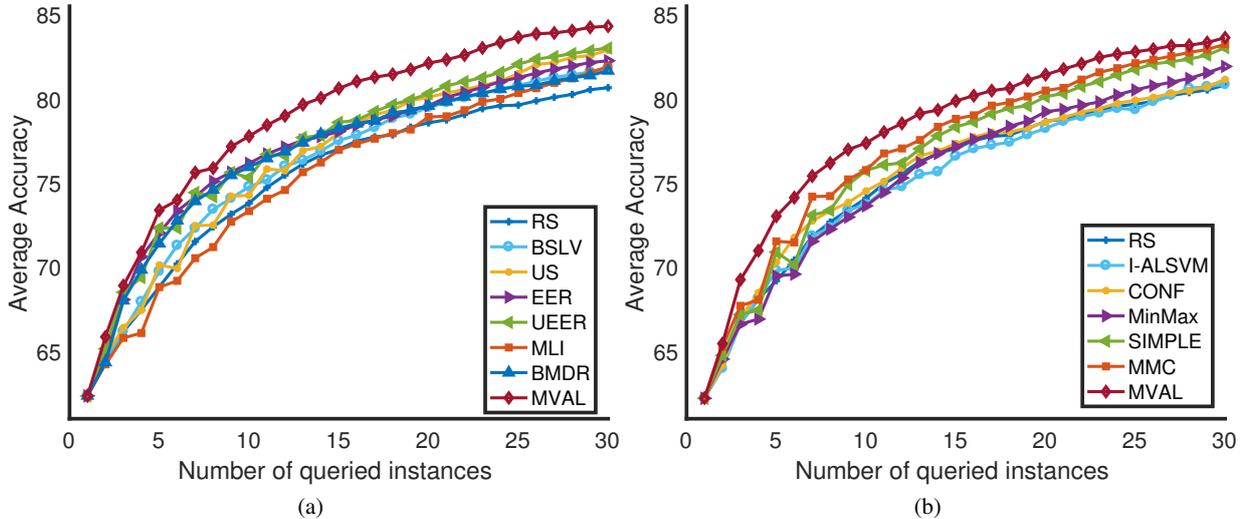

*Figure 3.* Average accuracy of the first 30 labeled examples over all the datasets. (a) shows the average performance of active learning methods based on logistic regression; (b) demonstrates the average result of active learning methods based on SVM.

- **CONF**: confidence-based active learning, which estimates the uncertainty by its conditional error (Li & Sethi, 2006);

- **I-ALSVM**: inconsistency-based active learning, which considers two extreme hypotheses and selects instance with highest inconsistency value (Wang et al., 2012);

- **MMC**: maximum model change, an adaptive version of simple margin. It selects the instance close to the decision boundary but also considers about its contribution to model change (Cai et al., 2014);

- **MinMax**: min-max view active learning, a new version of QUIRE (Huang et al., 2014), but uses the hinge loss instead;

For all the methods, we use linear SVM from the LIBSVM package (Chang & Lin, 2011) as classifier. The regularization parameter $C$ is set to 10 in all the experiments. As in the previous subsection, we use the area under the learning curve (ALC) as the performance measure. Like for the hyper-parameters of the base classifiers, there typically are no additional labeled validation data available for tuning any hyper-parameters an active learning scheme might have. We empirically tuned these parameters over all the datasets to globally good working choices. For CONF, an uncertain threshold $c$ and bin size $nBin$ are needed. The resulting parameters we found were $c = 0.5$ and $nBin = 4$. For MMC, a hyper-parameter $\gamma$ is used to filter the instances within the margin. We validated this value from a candidate set and selected the one which presents the best overall performance. Finally, we set $\gamma = 0.01$ in our experiment. For I-ALSVM, we used the modified version of I-ALSVM which outperforms original I-ALSVM by combining I-ALSVM and the simple margin method (Wang et al., 2012). The modified version first selects a small candidate set based on original I-ALSVM, then chooses the instance which is closest to the decision boundary from the above subset. We tuned the size of this subset and finally this parameter was set 16 in our experiment. We note that the proposed method MVAL does not need to tune additional hyper-parameters.

As shown on the right side of Table 5, the proposed method also in this setting achieves the overall best performance. MVAL obtains 0.834 in terms of average ALC and performs best or at the best level on 31 datasets. The second best algorithm, MMC, only performs well on 16 datasets. Also here we used a paired $t$-test at 95% significance level to evaluate the scores of ALC over all 45 datasets and we can conclude that MVAL significantly outperforms other approaches. The win/tie/loss counts of MVAL versus other methods also demonstrate that MVAL compares favorably to all other methods. We also note that MMC and the simple margin outperform other active learning methods except MVAL. CONF and I-ALSVM perform slightly better than random sampling. The possible reason might be that their hyper-parameters need to be tuned very on each dataset. We plot the average performance of the first 30 annotated examples in Figure 3b. Also looking at the performance in this way, we see that MVAL performs better than other algorithms, especially in the early stage of active learning. MMC slightly outperforms SIMPLE and I-

*Table 5.* Performance comparisons of active learning algorithms in terms of the areas under the learning curve (ALC) based on SVM. "Average ALC" reports the average ALC scores over all the datasets. "Average Ranking" shows the average ranking within the compared methods. "Win Times" is the number of datasets on which an algorithm achieving the best or comparable performance. "W/T/L MVAL VS" represents the win/tie/loss counts of MVAL versus the other algorithms over all the datasets. Similarly, "W/T/L VS RS" shows the win/tie/loss counts of other methods versus random sampling.

| | RS | I-ALSVM | CONF | MinMax | SIMPLE | MMC | MVAL |
|---|---|---|---|---|---|---|---|
| hill | 0.534 | 0.549 | 0.550 | 0.578 | 0.580 | 0.581 | **0.587** |
| liver | 0.599 | **0.622** | 0.613 | 0.611 | 0.622 | 0.612 | 0.609 |
| cylinder | 0.602 | 0.592 | 0.603 | 0.631 | 0.608 | **0.637** | **0.641** |
| splice | 0.658 | 0.667 | 0.658 | 0.630 | **0.682** | 0.670 | 0.670 |
| religionatheism | 0.673 | **0.677** | 0.673 | 0.641 | 0.673 | 0.662 | 0.647 |
| ooctris2f | **0.682** | 0.664 | **0.681** | 0.638 | 0.662 | 0.654 | 0.680 |
| musk1 | 0.687 | 0.691 | 0.687 | 0.700 | 0.680 | 0.662 | **0.703** |
| pcmac | 0.690 | 0.679 | 0.690 | 0.674 | 0.680 | **0.711** | 0.675 |
| pima | 0.705 | 0.701 | 0.702 | **0.727** | 0.714 | **0.726** | 0.713 |
| sonar | 0.705 | 0.725 | 0.708 | 0.710 | 0.720 | 0.714 | **0.731** |
| diabetes | 0.715 | 0.687 | 0.717 | 0.726 | 0.720 | 0.722 | **0.736** |
| fertility | 0.729 | **0.752** | 0.733 | 0.738 | 0.757 | **0.760** | **0.752** |
| basehockey | 0.730 | 0.751 | 0.730 | 0.706 | 0.743 | **0.765** | 0.722 |
| blood | 0.732 | 0.725 | 0.732 | **0.747** | 0.726 | 0.735 | 0.740 |
| hepatitis | 0.750 | 0.759 | 0.755 | 0.760 | 0.776 | **0.779** | **0.771** |
| heart | 0.756 | 0.777 | 0.776 | 0.785 | 0.775 | 0.783 | **0.790** |
| baseball | 0.768 | 0.838 | 0.766 | 0.842 | 0.850 | **0.867** | 0.859 |
| autos | 0.771 | 0.836 | 0.775 | 0.844 | 0.857 | 0.852 | **0.869** |
| motorcycles | 0.776 | 0.849 | 0.777 | 0.861 | 0.862 | 0.870 | **0.884** |
| mammographic | 0.784 | 0.768 | 0.790 | 0.782 | 0.795 | **0.803** | 0.791 |
| ionosphere | 0.791 | 0.779 | 0.793 | 0.694 | 0.796 | 0.793 | **0.811** |
| australian | 0.793 | 0.801 | 0.819 | **0.844** | 0.835 | 0.832 | 0.818 |
| hockey | 0.798 | 0.878 | 0.797 | 0.888 | 0.880 | **0.899** | **0.898** |
| vc2 | 0.803 | 0.779 | 0.814 | 0.822 | 0.793 | 0.811 | **0.828** |
| parkinsons | 0.824 | 0.832 | 0.829 | 0.835 | **0.845** | **0.845** | 0.835 |
| letterIJ | 0.847 | 0.787 | 0.863 | 0.867 | 0.868 | 0.879 | **0.891** |
| vehicle | 0.857 | 0.845 | 0.864 | 0.877 | 0.881 | 0.877 | **0.887** |
| 7vs9 | 0.858 | 0.883 | 0.869 | 0.900 | 0.901 | 0.907 | **0.918** |
| 5vs8 | 0.859 | 0.883 | 0.876 | 0.854 | 0.891 | 0.888 | **0.910** |
| letterVY | 0.860 | 0.778 | 0.867 | 0.856 | 0.868 | 0.876 | **0.882** |
| 3vs5 | 0.864 | 0.871 | 0.859 | 0.857 | 0.884 | 0.880 | **0.895** |
| glass | 0.897 | 0.903 | 0.895 | **0.907** | 0.902 | 0.903 | **0.909** |
| wine | 0.897 | 0.904 | 0.898 | 0.926 | 0.932 | **0.930** | **0.939** |
| tictactoe | 0.904 | 0.848 | **0.908** | 0.870 | 0.894 | 0.894 | 0.912 |
| letterMN | 0.912 | 0.872 | 0.912 | 0.927 | 0.934 | 0.934 | **0.947** |
| wdbc | 0.918 | 0.945 | 0.925 | 0.958 | 0.956 | 0.957 | **0.961** |
| letterEF | 0.926 | 0.921 | 0.927 | 0.956 | 0.956 | **0.959** | **0.960** |
| ozone | 0.928 | 0.942 | 0.928 | 0.930 | 0.937 | 0.934 | **0.945** |
| mushrooms | 0.931 | 0.968 | 0.930 | 0.964 | 0.970 | 0.970 | **0.973** |
| letterDP | 0.935 | 0.917 | 0.935 | 0.964 | 0.959 | 0.964 | **0.967** |
| ac-inflam | 0.942 | 0.943 | 0.942 | **0.977** | **0.979** | **0.979** | **0.975** |
| wisc | 0.944 | 0.936 | 0.940 | **0.953** | 0.951 | **0.953** | 0.950 |
| breast | 0.947 | 0.955 | 0.956 | **0.964** | 0.963 | 0.963 | 0.963 |
| acute | 0.949 | 0.955 | 0.949 | 0.996 | **0.988** | **0.989** | **0.984** |
| letterUV | 0.949 | 0.939 | 0.950 | **0.976** | 0.974 | **0.977** | **0.979** |
| Average ALC | 0.804 | 0.808 | 0.808 | 0.819 | 0.827 | 0.830 | **0.834** |
| Average Ranking | 5.64 | 5.11 | 5.22 | 4.04 | 3.16 | 2.69 | **2.13** |
| Win Times | 1 | 3 | 2 | 9 | 5 | 16 | **31** |
| W/T/L MVAL VS | 41/0/4 | 39/2/4 | 38/3/4 | 32/8/5 | 31/6/8 | 26/10/9 | - |
| W/T/L VS RS | - | 24/3/18 | 18/23/4 | 30/5/10 | 38/4/3 | 38/4/3 | **41/0/4** |

ALSVM performs similarly to random sampling.

# 6. Experiments with Multi-Class Classification

We present the experimental results on multi-class classification problems in this section. Since many of the compared active learning algorithms using SVM are only designed for binary case and it is not clear how to extend them to multi-class problems, we only compare the proposed method with active learning algorithms that are derived on the basis of logistic regression.

We use 12 UCI benchmark datasets and 8 real-word datasets as the test bed. For some relatively large datasets such as MNIST, scene13 (Fei-Fei & Perona, 2005), GT-SRB (Stallkamp et al., 2011) and CIFAR10 (Krizhevsky & Hinton, 2009), we use randomly sub-sampling to reduce their sizes. The datasets information after sub-sampling and PCA is listed in Table 6. For the scene13 dataset, we use the GIST feature (Oliva & Torralba, 2001); for the CIFAR10 dataset and GTSRB dataset, HOG feature (Dalal & Triggs, 2005) are extracted. With regards to the action recognition datasets, KTH (Schuldt et al., 2004) and UCFsports (Rodriguez et al., 2008), we use the pre-extracted Action Bank features (Sadanand & Corso, 2012). The Isolet is a letter speech recognition dataset (Fanty & Cole, 1991). TWSA03 is a player action recognition data set in tennis games taken from (De Campos et al., 2011), of which HOG3D descriptors are extracted according to (Klaser et al., 2008).

*Table 6.* Multi-class datasets information after pre-processing: the number of instances (# Ins), the feature dimensionality (# Fea) and class number (#C)

| Dataset (#Ins, #Fea, #C) | Dataset (#Ins, #Fea, #C) | Dataset (#Ins, #Fea, #C) |
|---|---|---|
| car (900, 6, 4) | led.display (1000, 7, 10) | heart_cleveland(303, 13, 5) |
| contrac (1473, 9, 3) | pendigits (1000, 16, 10) | satimage (1000, 36, 6) |
| segment (1000, 19, 7) | stvehicle (846, 18, 4) | glass (214, 9, 6) |
| dermatology (366, 34, 6) | vowel (990, 10, 11) | USPS (1000, 60, 10) |
| MNIST(1000, 60, 10) | scene13 (1000, 90,13) | CIFAR10 (1000, 57,10) |
| KTH (599, 100, 6) | UCFsports(140, 100,10) | TWSA03 (1228,100, 3) |
| GTSRB (1000, 40, 20) | Isolet(1040, 40, 26) | |

The experiments are repeated 10 times on each datasets and average performances are reported. As the initial training set, we randomly select one instance from each class. For the logistic regression classifier, the same setting is used as that in Section 5.3. Due to that BSLV, Logistic Bound, BMDR are specifically designed for binary tasks, they are omitted for comparison. The proposed method MVAL is compared with the remaining active learning algorithms.

As is shown in Table 7, MVAL consistently outperforms other active learning methods over 14 datasets, it achieves the best performance or behaves comparably to the best algorithms. Though it fails on 6 datasets such as CIFAR10, MNIST and dermatology, it is never the worst one. This can demonstrate the advantages of MVAL, efficient and robust. We also observe that MLI totally fails on most of datasets and performs worse than random sampling. The probable reason may be that the min-max view used in (Huang et al., 2014) is not suitable for multi-class classification problems. The error reduction method EER achieves the second best scores while MVAL still outperforms it on 17 datasets based on paired $t$-test at a 95% significance level. Three variance-minimization approaches, $V$-opt, $\Sigma$-opt and TED, perform better than random sampling. How-ever, they are still worse than the proposed method, e.g. the win/tie/loss of MVAL versus $\Sigma$-opt is $17/1/2$.

*Table 7.* Performance comparisons of active learning algorithms on 20 multiclass datasets. "Average ALC" reports the average ALC scores over all the datasets. "Average Ranking" shows the average ranking in the compared methods. "Win Times" is the number of datasets on which an algorithm achieving the best or comparable performance. "W/T/L MVAL VS" represents the win/tie/loss counts of MVAL versus the other algorithms over all the datasets. Similarly, "W/T/L VS RS" shows the win/tie/loss counts of other methods versus random sampling.

| | RS | US | EER | UEER | MLI | $V$-opt | $\Sigma$-opt | TED | MVAL |
|---|---|---|---|---|---|---|---|---|---|
| CIFAR10 | 0.257 | 0.253 | **0.270** | 0.261 | 0.240 | 0.249 | **0.269** | 0.253 | 0.256 |
| vowel | 0.378 | 0.374 | 0.391 | 0.388 | 0.373 | 0.401 | 0.385 | 0.381 | **0.413** |
| contrac | 0.434 | **0.444** | 0.440 | 0.437 | 0.393 | 0.443 | 0.441 | **0.446** | 0.443 |
| scene13 | 0.471 | 0.418 | 0.500 | 0.443 | 0.420 | 0.487 | 0.465 | 0.476 | **0.504** |
| heart_cleveland | 0.501 | 0.522 | 0.521 | 0.527 | 0.514 | 0.507 | 0.517 | 0.512 | **0.531** |
| glass | 0.521 | **0.542** | 0.535 | 0.526 | 0.520 | 0.491 | **0.549** | 0.473 | 0.539 |
| GTSRB | 0.628 | 0.621 | 0.669 | 0.644 | 0.674 | 0.643 | 0.664 | 0.677 | **0.681** |
| MNIST | 0.628 | 0.627 | **0.709** | 0.649 | 0.587 | 0.685 | 0.692 | 0.674 | 0.700 |
| Isolet | 0.629 | 0.631 | 0.645 | 0.631 | 0.637 | 0.592 | 0.651 | 0.654 | **0.659** |
| led_display | 0.633 | **0.662** | 0.653 | 0.659 | 0.542 | 0.640 | 0.641 | 0.632 | **0.663** |
| stvehicle | 0.652 | 0.664 | 0.668 | 0.675 | 0.631 | 0.643 | 0.659 | 0.662 | **0.680** |
| car | 0.694 | 0.730 | 0.725 | **0.735** | 0.627 | 0.727 | 0.729 | 0.710 | 0.734 |
| pendigits | 0.752 | 0.766 | 0.770 | 0.760 | 0.733 | 0.734 | 0.735 | 0.766 | **0.786** |
| satimage | 0.763 | 0.746 | 0.766 | 0.764 | 0.760 | 0.759 | 0.753 | 0.745 | **0.793** |
| USPS | 0.769 | 0.797 | **0.816** | 0.802 | 0.777 | 0.804 | 0.812 | 0.798 | **0.817** |
| UCFsports | 0.769 | 0.769 | 0.758 | 0.770 | 0.788 | 0.766 | 0.766 | **0.797** | 0.775 |
| TWSA03 | 0.775 | 0.795 | 0.789 | 0.803 | 0.787 | 0.764 | 0.799 | 0.794 | **0.811** |
| segment | 0.809 | 0.810 | 0.828 | 0.850 | 0.794 | 0.842 | 0.846 | 0.820 | **0.865** |
| KTH | 0.918 | 0.951 | 0.936 | 0.948 | 0.932 | 0.927 | 0.942 | 0.941 | **0.953** |
| dermatology | 0.940 | 0.945 | 0.940 | **0.952** | 0.936 | 0.913 | 0.925 | 0.939 | 0.950 |
| Average ALC | 0.646 | 0.653 | 0.667 | 0.661 | 0.633 | 0.651 | 0.661 | 0.661 | **0.678** |
| Average Ranking | 6.9 | 5.15 | 3.85 | 3.85 | 7.25 | 6.3 | 4.85 | 5.2 | **1.65** |
| Win Times | 0 | 3 | 3 | 2 | 0 | 1 | 3 | 2 | **14** |
| W/T/L MVAL VS | 19/1/0 | 16/4/0 | 17/1/2 | 17/1/2 | 19/0/1 | 19/1/0 | 17/1/2 | 18/0/2 | - |
| W/T/L VS RS | - | 12/3/5 | 18/1/1 | 17/2/1 | 7/3/10 | 10/1/9 | 15/0/5 | 14/3/3 | **19/1/0** |

## 7. Discussion and Conclusion

We proposed a novel active learning method called MVAL, which is based on the retraining-based active learning framework. MVAL builds weighted retraining information matrices (RIMs) to record the changes of the output of unlabeled data during the retraining process. Two types of variance based on these RIMs are calculated and fused to evaluate the combined informativeness and representativeness of unlabeled samples. MVAL then selects the instance with the largest combined variance. As an example, we demonstrated how to use MVAL both with logistic regression and support vector machines. Furthermore, an extension of MVAL to multi-class classification task is also presented in this paper. Empirical results on both binary and multi-class datasets show excellent performance of our method in comparison with current state-of-the-art active learning methods.

We see two different extension of our approach as potentially interesting for future research. First of all, currently, MVAL is only feasible for myopic active learning setting. Like for many other active learning approaches, it may be interesting to investigate how to extend this idea to batch mode active learning, which queries a set of unlabeled examples simultaneously. Secondly, if there is one drawback our method has, it is the computational cost. It is not a problem that only our method has: MVAL actually has the same computational complexity as some of the state-of-the-art retraining-based methods that we compared to, namely EER (Roy & Mccallum, 2001), UEER (Yang & Loog, 2016), and MLI (Huang et al., 2014). For some simple active learning methods, such as uncertainty sampling, a proper acceleration can be achieved by hyperplane hashing (Liu et al., 2012). For our method and other retraining-based approaches, a feasible solution is to use parallel computing to improve the efficiency since retraining the classifier with different $x_i \in \mathcal{U}$ is independent of each other. Another direction to speed up these methods is using various heuristic approximations (e.g. a warm start in (Guo & Greiner, 2007) and nearly zero assumption of the gradient of objective function in (Settles & Craven, 2008)) and subsampling strategies (e.g. selecting a subset of samples with maximum entropy (Wei et al., 2015)).

More important than the extension to the batch setting and the computational speed is that we at all have a criterion that can give us good active learning performance. With the current work, we have made an additional step in this direction, clearly improving upon current state of the art.

# Appendix

*Table A.1.* Performance comparisons of the proposed method versus variance-minimization algorithms on 41 binary datasets.

| | RS | V-opt | Σ-opt | TED | Logistic Bound | MVAL |
|---|---|---|---|---|---|---|
| hill | 0.583 | 0.600 | 0.585 | 0.592 | 0.588 | **0.621** |
| cylinder | 0.596 | 0.583 | **0.612** | 0.598 | 0.603 | 0.602 |
| liver | **0.628** | **0.630** | 0.623 | 0.621 | 0.625 | **0.631** |
| splice | 0.651 | 0.612 | 0.667 | 0.661 | **0.671** | **0.671** |
| ooctris2f | 0.686 | 0.608 | 0.647 | 0.682 | 0.662 | **0.699** |
| musk1 | 0.688 | 0.658 | 0.684 | 0.676 | 0.692 | **0.713** |
| sonar | 0.688 | 0.661 | 0.669 | **0.707** | 0.704 | **0.704** |
| pcmac | 0.688 | **0.730** | 0.687 | 0.679 | 0.712 | 0.724 |
| religionatheism | 0.689 | 0.691 | **0.708** | 0.625 | 0.638 | 0.698 |
| fertility | 0.694 | 0.698 | 0.695 | 0.671 | 0.693 | **0.717** |
| pima | 0.704 | 0.699 | 0.716 | 0.676 | **0.729** | 0.725 |
| diabetes | 0.708 | 0.715 | 0.715 | 0.702 | **0.732** | 0.731 |
| blood | 0.727 | **0.736** | **0.736** | 0.701 | 0.732 | **0.739** |
| hepatitis | 0.727 | 0.655 | 0.695 | 0.721 | 0.746 | **0.767** |
| heart | 0.758 | 0.777 | 0.770 | 0.776 | 0.787 | **0.791** |
| basehockey | 0.766 | 0.815 | 0.737 | 0.733 | 0.785 | **0.822** |
| mammographic | 0.783 | 0.778 | 0.785 | 0.767 | 0.780 | **0.793** |
| australian | 0.785 | 0.802 | 0.802 | 0.806 | 0.830 | **0.848** |
| ionosphere | 0.797 | 0.720 | 0.798 | 0.802 | **0.820** | **0.822** |
| parkinsons | 0.811 | 0.823 | 0.815 | 0.809 | **0.827** | 0.825 |
| vc2 | 0.812 | 0.811 | 0.818 | 0.812 | **0.822** | 0.814 |
| letterIJ | 0.849 | 0.871 | 0.872 | 0.846 | 0.887 | **0.891** |
| 5vs8 | 0.855 | 0.892 | 0.907 | 0.905 | **0.915** | **0.914** |
| 7vs9 | 0.856 | 0.903 | 0.906 | 0.903 | **0.919** | **0.919** |
| vehicle | 0.858 | 0.885 | 0.875 | 0.877 | 0.896 | **0.900** |
| letterVY | 0.864 | 0.872 | 0.885 | 0.846 | 0.885 | **0.893** |
| 3vs5 | 0.866 | 0.880 | 0.896 | 0.887 | **0.904** | 0.902 |
| ozone | 0.877 | 0.635 | 0.827 | 0.859 | **0.900** | 0.887 |
| tictactoe | 0.896 | 0.865 | **0.922** | 0.899 | 0.889 | 0.905 |
| wine | 0.899 | 0.930 | 0.918 | 0.908 | 0.944 | **0.948** |
| glass | 0.899 | **0.911** | 0.909 | 0.905 | 0.911 | **0.913** |
| letterMN | 0.911 | 0.933 | 0.936 | 0.932 | **0.945** | **0.944** |
| wdbc | 0.916 | 0.938 | 0.933 | 0.932 | **0.955** | **0.954** |
| mushrooms | 0.931 | 0.965 | 0.966 | 0.961 | 0.975 | **0.976** |
| letterEF | 0.934 | 0.945 | 0.940 | 0.939 | **0.962** | **0.962** |
| letterDP | 0.938 | 0.954 | 0.953 | 0.957 | 0.968 | **0.970** |
| breast | 0.943 | 0.962 | 0.954 | 0.962 | **0.964** | 0.961 |
| ac-inflam | 0.947 | **0.982** | **0.980** | 0.974 | **0.983** | **0.981** |
| wisc | 0.949 | 0.953 | 0.948 | **0.955** | **0.956** | 0.956 |
| letterUV | 0.949 | 0.956 | 0.953 | 0.967 | **0.976** | **0.978** |
| acute | 0.967 | **0.990** | **0.991** | **0.990** | **0.990** | **0.991** |
| Average ALC | 0.807 | 0.806 | 0.815 | 0.810 | 0.827 | **0.834** |
| Average Ranking | 4.93 | 4.05 | 3.63 | 4.49 | 2.29 | **1.61** |
| Win Times | 1 | 6 | 6 | 3 | 18 | **30** |
| W/T/L MVAL VS | 40/1/0 | 33/7/1 | 34/3/4 | 34/6/1 | 21/13/7 | - |
| W/T/L VS RS | - | 27/3/11 | 28/6/7 | 20/8/13 | 35/2/4 | **40/1/0** |